\documentclass[letterpaper]{article} 
\usepackage{aaai25}  
\usepackage{times}  
\usepackage{helvet}  
\usepackage{courier}  
\usepackage[hyphens]{url}  
\usepackage{graphicx} 
\usepackage{natbib}  
\usepackage{caption} 
\usepackage{algorithm}
\usepackage{algorithmic}

\usepackage{amsmath}
\usepackage{mdframed}
\usepackage{multirow}
\usepackage{xcolor}
\usepackage{newfloat}
\usepackage{listings}
\DeclareCaptionStyle{ruled}{labelfont=normalfont,labelsep=colon,strut=off} 
\floatstyle{ruled}
\newfloat{listing}{tb}{lst}{}
\floatname{listing}{Listing}
\title{Guiding Exploration in Reinforcement Learning Through LLM-Augmented Observations}
\author {
    Vaibhav Jain\textsuperscript{\rm 1,2},
    Gerrit Grossmann\textsuperscript{\rm 1}
}
\affiliations {
    \textsuperscript{\rm 1}Deutsches Forschungszentrum für Künstliche Intelligenz\\
    \textsuperscript{\rm 2}Saarland University\\
    vaibhav.jain@dfki.de, gerrit.grossmann@dfki.de
}

\begin{document}

\maketitle

\begin{abstract}
Reinforcement Learning (RL) agents often struggle in sparse-reward environments where traditional exploration strategies fail to discover effective action sequences. Large Language Models (LLMs) possess procedural knowledge and reasoning capabilities from text pretraining that could guide RL exploration, but existing approaches create rigid dependencies where RL policies must follow LLM suggestions or incorporate them directly into reward functions. We propose a framework that provides LLM-generated action recommendations through augmented observation spaces, allowing RL agents to learn when to follow or ignore this guidance. Our method leverages LLMs' world knowledge and reasoning abilities while maintaining flexibility through soft constraints. We evaluate our approach on three BabyAI environments of increasing complexity and show that the benefits of LLM guidance scale with task difficulty. In the most challenging environment, we achieve 71\% relative improvement in final success rates over baseline. The approach provides substantial sample efficiency gains, with agents reaching performance thresholds up to 9 times faster, and requires no modifications to existing RL algorithms. Our results demonstrate an effective method for leveraging LLM planning capabilities to accelerate RL training in challenging environments.
\end{abstract}

%

\section{Introduction}

Reinforcement learning agents often struggle in sparse-reward environments where successful actions are rare and difficult to discover through random exploration. This challenge becomes particularly pronounced in long-horizon tasks where agents must execute many correct sequential actions before receiving feedback. While traditional RL algorithms rely on exploration strategies that can be inefficient when only specific action sequences lead to success, classical planning methods require explicit world models that are challenging to construct. Recent advances in LLMs have demonstrated remarkable capabilities in sequential decision-making \cite{kojima2022large}, leveraging their extensive pre-training on text that includes procedural knowledge and causal reasoning patterns absent in typical RL training. This contrasts with classical RL methods that must discover these patterns through trial-and-error exploration. This complementary strength raises the question: can LLMs provide effective planning guidance for RL agents in sparse reward settings?

Previous research has explored several approaches to combining LLMs with RL, including using LLMs as reward models \cite{du2023guiding,kwon2023reward}, hierarchical planners \cite{ahn2022can}, and direct action policies \cite{li2022pre}. However, these approaches share a fundamental limitation: they create hard constraints that require LLM guidance to be consistently accurate. Hard constraints in this context refer to system designs where RL policies heavily rely on LLM suggestions through hierarchical execution or direct policy control, or where LLM outputs directly modify reward functions and policy learning objectives. When LLM guidance is incorrect—due to limited environmental understanding or reasoning errors—these hard constraint methods can degrade agent performance since the RL agent has limited control over selectively using or ignoring suggestions.

We propose an approach that integrates LLM planning guidance through enhanced observations rather than direct action control. Throughout this paper, we use the terms LLM hints, LLM suggestions, and LLM guidance interchangeably to refer to next action or subgoal recommendations generated by the language model to assist RL exploration. The key insight is to give RL agents access to these LLM suggestions as additional observational input, allowing agents to learn whether to follow or ignore the guidance during training. This creates a collaborative system with soft constraints where LLMs provide planning hints while RL agents maintain full control over final decisions. Agents receive LLM hints at regular intervals and learn to incorporate this information alongside environmental observations through standard policy learning. When LLM suggestions are helpful, agents learn to follow them; when suggestions are incorrect, agents learn to ignore them. This approach works with any standard RL algorithm without requiring modifications to training procedures or specialized hierarchical components.

Our main contributions are: (1) We introduce a method for integrating LLM planning guidance into RL training through enhanced observations that creates soft rather than hard constraints. (2) We demonstrate that LLM guidance provides greater benefits in more complex environments, achieving up to 71\% relative improvement in success rates. (3) We show significant improvements in learning speed, with agents reaching performance thresholds up to 9 times faster.

\section{Related Work}

\paragraph{LLMs in Reinforcement Learning.} 
Recent work has explored several approaches for integrating LLMs into reinforcement learning systems. These include using LLMs as information processors to enhance observations with implicit knowledge \cite{paischer2022history, paischer2023semantic}, as reward models that evaluate trajectories and provide reward signals \cite{kwon2023reward, du2023guiding, li2024auto, ma2023eureka}, as planners that generate high-level subgoal sequences for hierarchical execution \cite{ahn2022can,carta2023grounding}, and as policies that directly assist with action selection \cite{li2022pre}. While these approaches have shown promise, they typically create hard constraints that require LLM suggestions to be consistently accurate, making them vulnerable to LLM errors and limiting agent learning flexibility.

\paragraph{LLMs for Planning.}
Large language models have shown promise for planning tasks, particularly in \emph{commonsense} reasoning and generating high-level plans. However, LLMs alone often struggle with long-horizon or complex planning, producing incomplete or non-executable plans \cite{Valmeekam2023On, Huang2024Understanding,Wei2025PlanGenLLMs:}. To address these limitations, hybrid approaches combine LLMs with classical planners, where LLMs provide problem translation and heuristic guidance while symbolic planners ensure correctness \cite{Liu2023LLM+P:,Sharan2023LLM-Assist:,Zhou2023ISR-LLM:,Kambhampati2024LLMs}. These hybrid methods demonstrate that LLMs work best when providing guidance rather than direct control, similar to our approach of using LLM suggestions as soft constraints in RL training \cite{Song2022LLM-Planner:,Zhou2023ISR-LLM:,Dagan2023Dynamic}.

\section{LLM-Guided Reinforcement Learning}
\subsection{Problem Formulation}

We consider partially observed Markov decision processes (POMDPs) defined by the tuple $(S, A, O, \Omega, T, \gamma, R)$, where observations $o \in \Omega$ are generated from environment states $s \in S$ and actions $a \in A$ via the observation function $O(o \mid s, a)$. The transition function $T(s' \mid s, a)$ describes the environment dynamics, while $R$ and $\gamma$ represent the reward function and discount factor, respectively.

Our approach operates within a goal-conditioned RL setting, where the agent is trained to reach a specified goal state $g \in S$ from an initial state $s_0 \in S$. The key innovation lies in enhancing the observation space by incorporating LLM suggestions to create enriched observations $o' \in \Omega'$, where the hint space includes both the hint value and an availability indicator.

Formally, we define the enhanced observation as:
$$o' = \{o, h, h_{\text{avail}}\}$$
where $o$ represents the standard environmental observation, $h$ is the LLM-generated planning hint, and $h_{avail}$ is a binary flag indicating hint availability. This formulation allows the RL agent to access LLM guidance as part of its observational input while maintaining the flexibility to learn when this guidance should be utilized.

\subsection{LLM Hint Generation}

To generate LLM hints, we prompt the LLM with three key inputs: the encoded current environment state $\text{encode}(s)$, the last $p$ previous actions $a_{t-p:t-1}$, and the task description $g$. The hint generation function is formally defined as:
$$h = \text{LLM}(\text{encode}(s), a_{t-p:t-1}, g)$$

\begin{figure}[t]
    \centering
    \includegraphics[width=0.95\columnwidth]{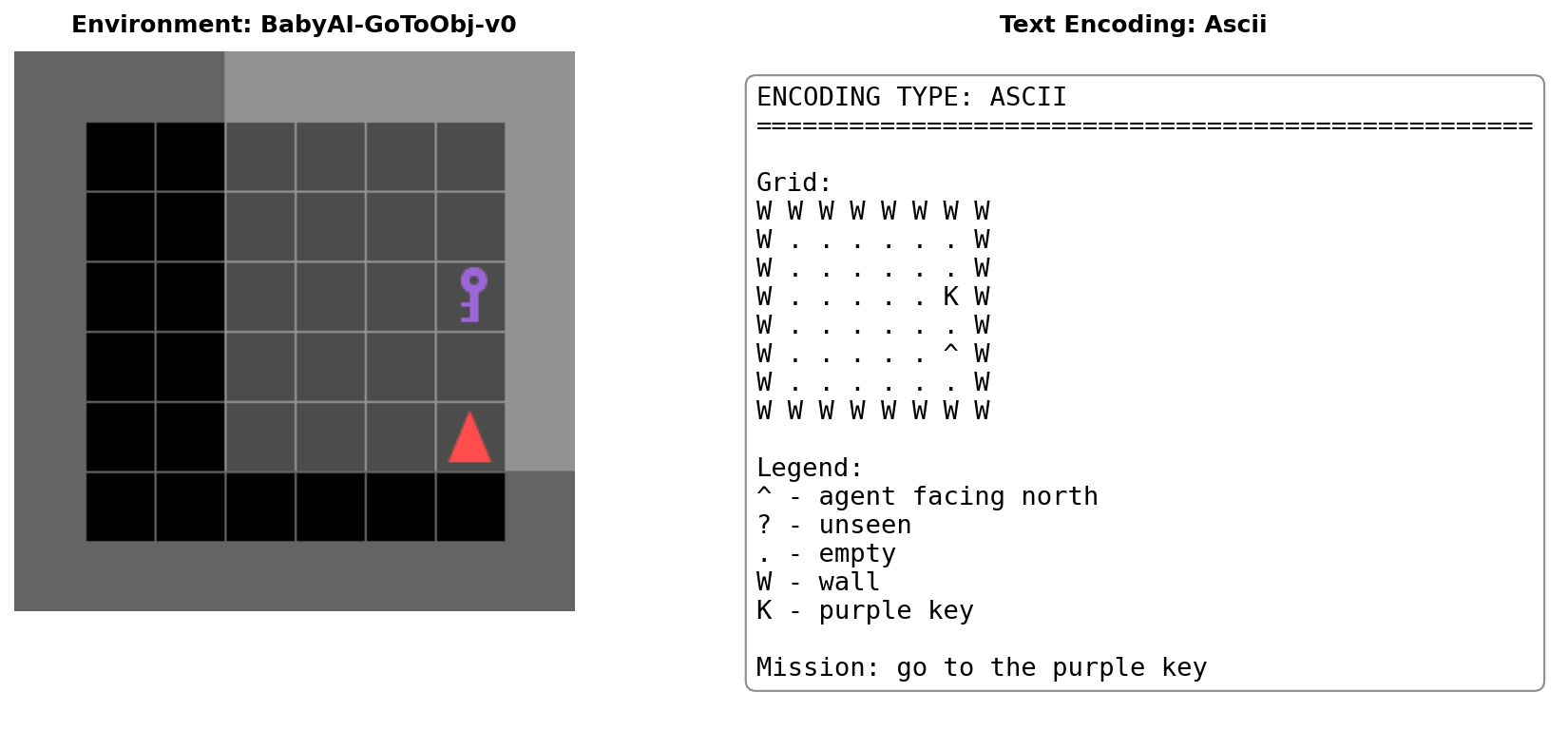}
    
    \vspace{0.3em}
    
    \begin{mdframed}[
        linewidth=1pt,
        linecolor=gray!50,
        backgroundcolor=white!5,
        roundcorner=3pt,
        innertopmargin=4pt,
        innerbottommargin=4pt,
        innerleftmargin=3pt,
        innerrightmargin=3pt
    ]
    \vspace{0.2em}
    
    {\scriptsize\ttfamily
    \begin{flushleft}
        \textbf{(Example) LLM Chain-of-Thought Response:}\\
        Prediction(\\
        \hspace*{1em}reasoning="The agent's mission is to go to the purple key. The purple key is located to the north of the agent. Since the agent is facing north, it first needs to move forward towards the purple key. The action to move the agent forward would be the most efficient first step.",\\
        \hspace*{1em}primitive\_action=2,\\
        \hspace*{1em}subgoal=GoNextToSubgoal\\
        )
        \end{flushleft}}
    \end{mdframed}
    
    \caption{Environment visualization with ASCII encoding (top) and corresponding LLM reasoning chain-of-thought response (bottom). The LLM analyzes the spatial relationships and mission requirements to determine the optimal action.}
    \label{fig:encoding_ascii}
\end{figure}

We encode environment states using grid-style ASCII maps that capture spatial relationships between the agent, objects, and environmental features in a structured textual format (Figure~\ref{fig:encoding_ascii}). We evaluated four different encoding methods and found ASCII encoding to be the most consistent and reliable for LLM reasoning (see Appendix for details). This representation provides the LLM with clear spatial context while remaining computationally efficient. Previous actions are maintained as a rolling history of the most recent $p$ actions, formatted as step-indexed strings to provide temporal context that helps the LLM understand the agent's recent behavior patterns.

The LLM processes this contextual information and generates structured hints containing both \emph{primitive actions and subgoals} (Figure~\ref{fig:encoding_ascii}, bottom). The primitive action integers are parsed from the LLM output and correspond directly to the RL agent's discrete action space.

During training, hints are generated every $k$ timesteps. When hints are provided, the agent receives its environmental observation, the LLM suggestion, and $h_{avail} = 1$. When no hints are provided, the agent receives a neutral hint value and $h_{avail} = 0$. This maintains consistent observation structure and allows the agent to learn when to use LLM guidance.

\begin{algorithm}[t]
\caption{LLM-Guided RL Training}
\begin{algorithmic}[1]
\STATE \textbf{Initialize} environment $env$, LLM model, RL agent $\pi$
\STATE \textbf{Set} hint frequency $k$, action history size $p$
\FOR{$\text{episode} = 1$ \textbf{to} $N$}
    \STATE $o_0, g \gets env.\text{reset()}$
    \STATE $a_{\text{history}} \gets [\,]$ // Initialize action history
    \STATE $t \gets 0$
    \STATE $\text{done} \gets \text{False}$
    \WHILE{\textbf{not} done}
        \STATE $t \gets t + 1$
        \IF{$t \bmod k = 0$}
            \STATE $s_{\text{encoded}} \gets \text{encode\_ascii}(o_t)$
            \STATE $h \gets \text{LLM}(s_{\text{encoded}}, a_{\text{history}}, g)$
            \STATE $h_{\text{avail}} \gets 1$
        \ELSE
            \STATE $h \gets \text{neutral\_action}$
            \STATE $h_{\text{avail}} \gets 0$
        \ENDIF
        \STATE $o'_t \gets \{o_t, h, h_{\text{avail}}\}$ // Enhanced observation
        \STATE $a_t \gets \pi(o'_t)$ // Agent action
        \STATE $o_{t+1}, r_t, \text{done} \gets env.\text{step}(a_t)$
        \STATE \textbf{Update} $a_{\text{history}}$ with $a_t$ // keep last $p$ actions
        \STATE \textbf{Train} agent $\pi$ using $(o'_t, a_t, r_t, o'_{t+1})$
    \ENDWHILE
\ENDFOR
\end{algorithmic}
\end{algorithm}

\subsection{Training Procedure}

The RL agent is trained using standard policy gradient methods on the enhanced observation space $\Omega'$. The agent learns a policy $\pi(a|o', h, h_{\text{avail}})$ that maps environmental observations, hint values, and hint availability to actions. This creates a soft constraint mechanism where the agent learns when LLM suggestions are helpful versus when to rely on its own policy.

Since the enhanced observations simply add new input fields, no modifications to existing RL algorithms are required. The agent learns to weight LLM guidance against its own value estimates through normal exploration and exploitation. When hints are available and useful, the agent learns to follow them. When hints are unavailable or incorrect, the agent learns to ignore them and act independently. Algorithm~1 outlines the complete training procedure. For RL training, we use PPO with standard hyperparameters \cite{schulman2017proximal}, training for 3M frames on simpler environments and 5M frames on complex environments.

\section{Experiments}

We perform experiments to test the following hypothesis: 
\begin{itemize}
    \item H1: LLM can act as useful planner proxy to provide context aware and sensible planning hints.
    \item H2: These LLM-generated planning hints can improve learning speed and performance in sparse reward environments.
\end{itemize}

\begin{table}[t]
    \centering
    \begin{tabular}{p{2.2cm} p{5.2cm}}
    \hline
    \textbf{Method} & \textbf{Description} \\
    \hline
    Baseline & Standard PPO without hints \\
    \hline
    LLM-hints (ours) & PPO with LLM action suggestions via enhanced observations \\
    \hline
    Oracle & PPO with ground truth hints from optimal BabyAI planner \\
    \hline
    \end{tabular}
    \caption{Comparison of methods evaluated in our experiments.}
    \label{tab:methods}
\end{table}

\paragraph{Environment Details.} We test on three BabyAI environments of increasing complexity \cite{chevalier2018babyai}. \textbf{GoToObj (easy)} requires going to an object inside a single room with no doors and no distractors—a simple navigation task with direct line-of-sight to target. \textbf{OpenDoor (medium)} requires going to a door specified by color or location, where the door is always unlocked and in the current room, requiring object identification and navigation. \textbf{PickupLoc (hard)} requires picking up an object described by its location within a single room, demanding spatial reasoning, object identification, and precise navigation to the target location.

\subsection{Results}

\paragraph{Quality of LLM Planning Hints.} To answer H1, we study whether LLMs can provide useful planning guidance by manually analyzing a random sample of 30 LLM suggestions across our three environments. We evaluate each suggestion on three criteria: whether it matches the optimal action (Optimal Policy), whether the LLM correctly interprets the ASCII grid representation (State Awareness), and whether the suggested action follows logically from the LLM's stated understanding (Action Reasoning). 

The majority of LLM suggestions (73\%) match the optimal action, demonstrating that the LLM can serve as a reasonable proxy for near-optimal planning. Most suggestions (76\%) show correct state interpretation from the ASCII input, indicating that our encoding successfully conveys spatial relationships to the language model. When the LLM understands the state correctly, it provides sound action reasoning 84\% of the time. While not perfect, these results suggest that LLM hints contain sufficient signal to guide RL exploration, particularly given our soft constraint design that allows agents to learn when to ignore incorrect suggestions.

\begin{figure*}[t]
    \centering
    \includegraphics[width=\textwidth]{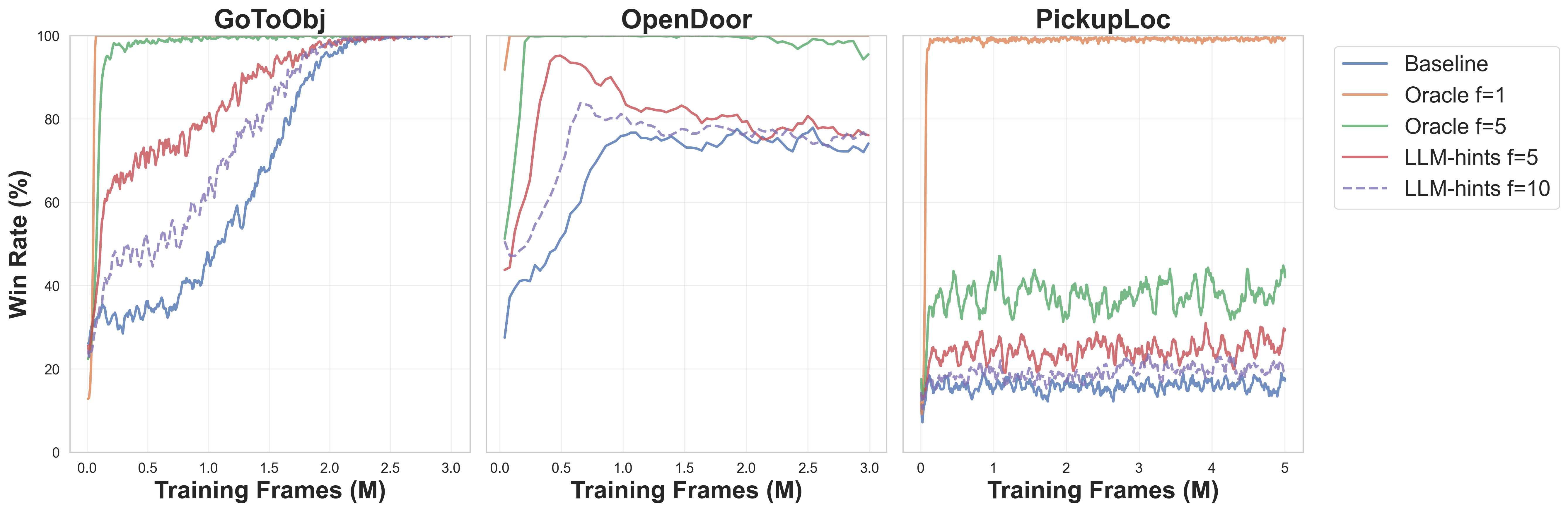}
    \caption{Training curves showing win-rate progression for frequency f=5 across environments. Our approach (LLM-hints) demonstrate faster convergence and higher final performance, particularly in complex environments like PickupLoc.}
    \label{fig:training_curves}
\end{figure*}

\begin{table*}[t]
    \centering
    \begin{tabular}{l c c c c c c}
    \hline
    \textbf{Environment} & \textbf{Baseline} & \textbf{+Text} & \textbf{LLM-hints f=5} & \textbf{LLM-hints f=10} & \textbf{LLM-hints +Text} & \textbf{Oracle f=5} \\
    \hline
    GoToObj & 100.0\% & 99.6\% & 100.0\% & 100.0\% & 100.0\% & 100.0\% \\
    OpenDoor & 74.2\% & 75.9\% & 75.0\% & 74.3\% & \textbf{77.9\%} & 96.4\% \\
    PickupLoc & 17.2\% & 15.9\% & \textbf{29.5\%} & 18.9\% & 20.4\% & 42.1\% \\
    \hline
    \end{tabular}
    \caption{Final win rates across environments for all evaluated methods and conditions. Columns include: Baseline, Baseline with mission text (+Text), LLM-hints at two frequencies (f=5, f=10), LLM-hints with mission text, and Oracle hints (f=5). The results show that benefits scale with task complexity, with the most dramatic improvements occurring in the hardest environment.}
    \label{tab:combined_performance}
\end{table*}

\paragraph{Impact on RL Training Performance.} For H2, we test whether LLM hints can improve both final performance and learning speed in sparse-reward environments. We train PPO agents with and without LLM hints across our three environments and measure final win rates after full training (Table~\ref{tab:combined_performance}) and sample efficiency to reach performance thresholds (Table~\ref{tab:sample_efficiency}).

LLM hints provide the greatest benefits in the most challenging environment, supporting our hypothesis that planning guidance becomes more valuable as task complexity increases. In PickupLoc, the hardest environment requiring spatial reasoning and precise navigation, agents with LLM hints achieve 29.5\% success compared to 17.2\% for baseline—a 71\% relative improvement. OpenDoor shows modest gains (74.2\% to 75.0\%), while GoToObj reaches ceiling performance for both conditions due to the task's simplicity. We hypothesize that PickupLoc benefits most because it requires long action sequences that are difficult to discover through random exploration, exactly where LLM planning guidance provides the most value.

The sample efficiency results reveal even more dramatic improvements. In GoToObj, agents with LLM hints reach 50\% success after only 120K frames compared to 1.08M frames for baseline—9× faster learning. In PickupLoc, baseline agents never reach 25\% success during training, while LLM-guided agents surpass this threshold and achieve 29.5\% final performance. These results demonstrate that LLM hints not only improve final performance but significantly accelerate the learning process by providing structured exploration guidance that helps agents discover successful behaviors more quickly.

\begin{table}[t]
    \centering
    \begin{tabular}{l c c c}
    \hline
    \textbf{Environment} & \textbf{Threshold} & \textbf{Baseline} & \textbf{LLM-hints f=5} \\
    \hline
    \multirow{2}{*}{GoToObj} & 50\% & 1.08M & \textbf{120K (9×)} \\
    & 75\% & 1.58M & \textbf{180K (8.8×)} \\
    \hline
    \multirow{2}{*}{OpenDoor} & 50\% & 475K & \textbf{400K (1.2×)} \\
    & 75\% & 950K & \textbf{800K (1.2×)} \\
    \hline
    PickupLoc & 25\% & Never & \textbf{Achieved} \\
    \hline
    \end{tabular}
    \caption{Sample efficiency: frames to reach success thresholds.}
    \label{tab:sample_efficiency}
    \end{table}

\paragraph{Hint Frequency Effects.} We test two hint frequencies across all environments: providing hints every 5 steps versus every 10 steps. The results show that k=5 consistently outperforms k=10 in all tested configurations (Table~\ref{tab:combined_performance}). This follows the natural hypothesis that more frequent guidance helps during early exploration phases when agents rely more heavily on external guidance to discover successful action sequences.

\section{Conclusion and Discussion}

We presented an approach for integrating LLM reasoning into RL training through enhanced observations that provide action hints as soft constraints. Our evaluation across three BabyAI environments demonstrates that LLM guidance provides benefits that scale with task complexity, achieving up to 71\% relative performance gains and 9× sample efficiency improvements in complex sparse-reward tasks.

The primary limitation is the computational cost of frequent LLM queries, which becomes prohibitive for large training runs. This makes the approach less suitable for domains requiring extensive training periods. Mitigation strategies include fine-tuning smaller LLMs, knowledge distillation, and model quantization. The soft constraint framework also enables adaptive hint scheduling, where hint frequency could be dynamically adjusted based on agent performance.

Exploring different hint abstractions and evaluating the approach on more complex environments and real-world robotics tasks are promising directions for future work.

\bibliography{aaai25}

\appendix

\section{Appendix}

\subsection{Implementation Details}

We use Llama3-70b with chain-of-thought prompting and grid-style ASCII state representations. We maintain a rolling history of the last $p=5$ previous actions. For RL training, we use PPO with standard hyperparameters, training for 3M frames on simpler environments and 5M frames on complex environments.

While our approach is algorithm-agnostic, we choose PPO for its stability in sparse-reward environments and widespread adoption as a standard benchmark. This enables clear isolation of LLM-hints impact. The enhanced observation framework can be adapted to other RL algorithms like SAC, A3C, or DQN variants without significant modification.

We evaluate hint frequencies of $k=5$ and $k=10$, with hints added as separate observation fields. We compare three PPO-based methods across our experiments: baseline PPO without hints, LLM-hints with enhanced observations, and Oracle-hints using ground truth suggestions from the optimal BabyAI planner.

\subsection{State Encoding Representations}

We evaluated four different methods for encoding environment states as text input to the LLM (Table~\ref{tab:encoding_subgoals}):

\begin{table}[t]
\centering
\begin{tabular}{l p{4.5cm}}
\hline
\multicolumn{2}{c}{\textbf{State Encoding Methods}} \\
\hline
\textbf{Encoding Type} & \textbf{Description} \\
\hline
Natural Language & Natural language descriptions with object positions \\
ASCII Grid & 2D symbolic grid with smart symbol mapping \\
Tuple List & Structured tuples with object properties \\
Relative Description & Position descriptions relative to agent orientation \\
\hline
\hline
\multicolumn{2}{c}{\textbf{Subgoal Types}} \\
\hline
\textbf{Subgoal} & \textbf{Description} \\
\hline
GoNextToSubgoal & Navigate next to a target object \\
PickupSubgoal & Pick up a specific object \\
DropSubgoal & Drop the currently carried object \\
OpenSubgoal & Open a door or container \\
CloseSubgoal & Close a door or container \\
ExploreSubgoal & Explore environment to locate objects \\
done & Task completed successfully \\
none & No specific subgoal (neutral state) \\
\hline
\end{tabular}
\caption{State encoding methods and subgoal types used in experiments.}
\label{tab:encoding_subgoals}
\end{table}

\paragraph{Natural Language.} Converts observations to natural language descriptions, listing objects and their positions in readable sentences. Example: "Agent is facing north. There is a red key at position (3,2). There is a blue door at position (5,4). Mission: go to the red key."

\paragraph{ASCII Grid.} Represents the environment as a 2D ASCII grid with symbolic notation for different objects and states. Objects are encoded using smart symbol mapping that adapts based on color combinations present in the environment. Example format shows agent direction with arrows (\textgreater, v, \textless, \textasciicircum) and includes a legend mapping symbols to object types.

\paragraph{Tuple List.} Encodes objects as structured tuples containing object type, color, optional state, and coordinates. Example: "Agent at (1,1) facing north. Objects: [('red' key, (3,2)), ('blue' door (open), (5,4))]. Mission: go to the red key."

\paragraph{Relative Description.} Describes object locations relative to the agent's current position and orientation, using directional terms like "ahead," "behind," "to your left," and "to your right." Includes distance information in Manhattan tiles.

Through systematic evaluation, we found ASCII grid encoding to be the most consistent and reliable representation for LLM reasoning. The ASCII format provides clear spatial relationships while maintaining compact representation, leading to more accurate action suggestions compared to other encoding methods.

\subsection{Subgoal vs. Primitive Action Hints}

\begin{figure}[t]
    \centering
    \includegraphics[width=\columnwidth]{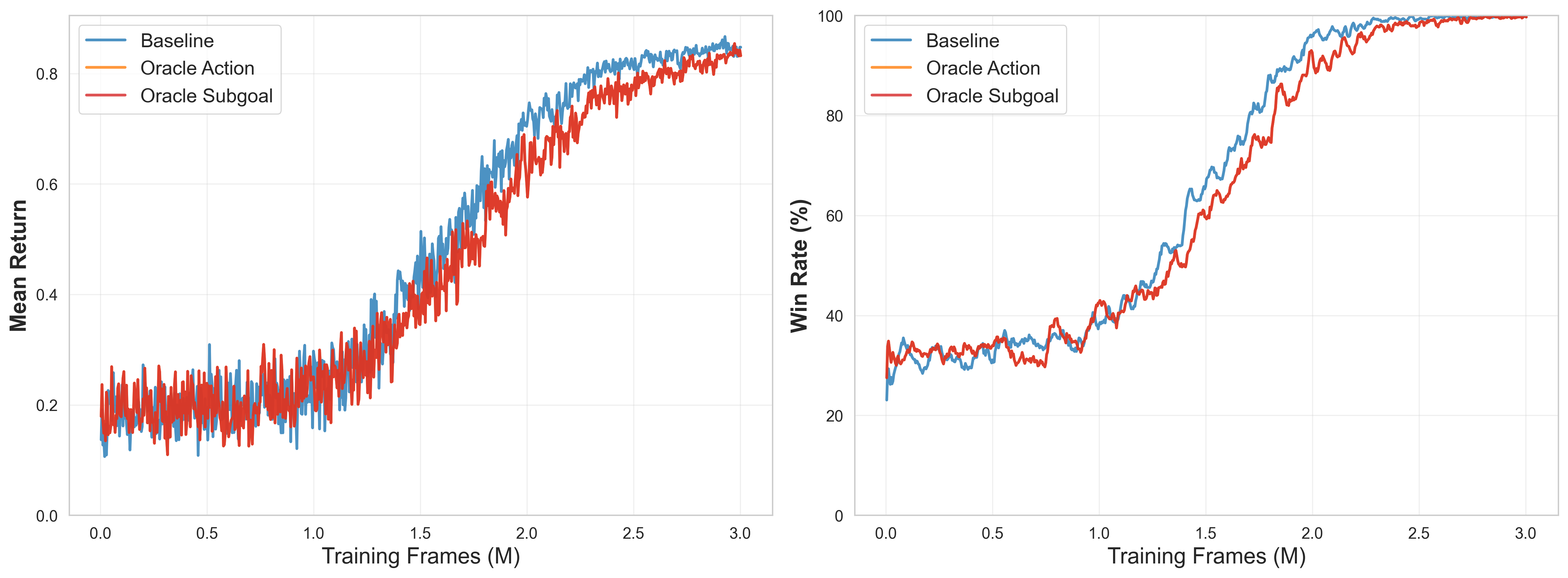}
    \caption{Training curves showing mean return and win-rate progression for GoToObj with Subgoal-based hints from Oracle. This illustrates that even with perfect subgoal hints, the agent still does show improvement over the baseline.}
    \label{fig:subgoal_comparison}
\end{figure}

We also experimented with subgoal hints but they did not perform as well as primitive action hints. The subgoal types used in our experiments are listed in Table~\ref{tab:encoding_subgoals}. While it is intuitive that low-level hints like "action-hints" would perform better than higher-level hints like "subgoals," our results show that subgoals do not seem to provide any improvement over baseline (Figure~\ref{fig:subgoal_comparison}). However, obtaining accurate and consistent low-level hints requires more prompt tuning and is less consistent than generating higher-level guidance. This suggests it may be worthwhile to investigate further the trade-offs between different levels of hint abstraction and their impact on RL performance.

In our implementation, subgoal hints represent strategic objectives that the agent should pursue, while primitive action hints provide direct action suggestions. These subgoals capture common behavioral patterns required for successful task completion, such as navigation (GoNextToSubgoal), object manipulation (PickupSubgoal, DropSubgoal), and environment interaction (OpenSubgoal, CloseSubgoal).

\end{document}